\documentclass[letterpaper, 10 pt, conference]{ieeeconf}
\usepackage{graphicx}
\usepackage{multirow}
\usepackage{amsmath}
\usepackage[caption=false]{subfig}
\usepackage{cleveref}
\usepackage{amsmath}
\usepackage{mathtools}
\usepackage{afterpage}
\usepackage[table,xcdraw]{xcolor}
\usepackage{graphicx}

\usepackage{lipsum}
\usepackage{multicol}
\usepackage{graphicx}

\IEEEoverridecommandlockouts                              

\title{\LARGE \bf
DroneLight: Drone Draws in the Air using Long Exposure Light Painting and ML
}

\author{Roman Ibrahimov$^{1*}$, Nikolay Zherdev$^{1*}$, and Dzmitry Tsetserukou$^{1}$
\thanks{$^{1}$Roman Ibrahimov, Nikolay Zherdev, and Dzmitry Tsetserukou are with the Intelligent Space Robotics Laboratory, Skolkovo Institute of Science and Technology, Moscow, Russian Federation.
        {\tt\small\{Roman.Ibrahimov, Nikolay.Zherdev, D.Tsetserukou\}@skoltech.ru}}%
\thanks{$^{*}$Authors contributed equally.}%
}

\begin{document}

\maketitle
\thispagestyle{empty}
\pagestyle{empty}

\begin{abstract}
We propose a novel human-drone interaction paradigm where a user directly interacts with a drone to light-paint predefined patterns or letters through hand gestures. The user wears a glove which is equipped with an IMU sensor to draw letters or patterns in the midair. The developed ML algorithm detects the drawn pattern and the drone light-paints each pattern in midair in the real time. The proposed classification model correctly predicts all of the input gestures. The DroneLight system can be applied in drone shows, advertisements, distant communication through text or pattern, rescue, and etc. To our knowledge, it would be the world’s first human-centric robotic system that people can use to send messages based on light-painting over distant locations (drone-based instant messaging). Another unique application of the system would be the development of vision-driven rescue system that reads light-painting by person who is in distress and triggers rescue alarm.
\vspace{10 mm}
\par
 
\end{abstract}

\section{Introduction}
Today, drones are becoming an integral part of our daily life.  The majority of the most demanding and high-paying jobs in many sectors, such as transport, agriculture, infrastructure, etc. are being gradually displaced by the emerging drone technology. In parallel, the global market of the business services that utilize drone technology is also gradually emerging. According to a study conducted by PricewaterhouseCoopers (PwC), the global market of the commercial applications of drones is estimated to be over 127 billion United States Dollars (USD) \cite{compton_2016}. Promising innovation, low costs, and higher efficiency of using such technology have also caught the attention of giant companies, like Amazon, FedEx, Uber, Microsoft, Google, Facebook. For example, Amazon, ever since 2013, has started a project for drone delivery to homes. The company’s Research and Development (R\&D) department has been working on its own new design of drones for delivering parcels to the customers within a short period of time. In 2016, the first Amazon parcel containing popcorn and TV stick was delivered by an autonomous drone in 13 minutes within a newly introduced service Prime Air.

\par The usage of drones rapidly increases and same holds for human interaction with drones. As a result, achieving natural human-drone interaction (HDI) techniques becomes an important challenge to address. When it comes controlling a single or multiple drones, a reliable, easy-to-learn, and tangible HDI is needed. Up to now, the available technologies have not made it possible to control drones intuitively without special training and additional control equipment. The aim of this work is to introduce a new way of interaction with a nano-quadcopter. The proposed system consists of a nano-quadcopter with LEDs, a glove that is equiped with an inertial measurement unit (IMU), and Machine Learning (ML) algorithm running on the base-station. The user wears the glove to draw the pre-defined patterns or letters in the air. The ML algorithm detects and matches each gesture to a unique letter or pattern. Then, the gestures are light-painted in the air by the nano-quadcopter as shown in Fig. 1. 
\
\par


\section{Related Work}
\par Being an interdisciplinary research field, Human-Robot Interaction (HRI) emerges from the intersection of multiple domains, such as Robotics, Artificial Intelligence, Engineering, Human-Computer Interaction, Cognitive Psychology, and Social Sciences \cite{dautenhahn2007methodology}. Building an explicit or implicit communication between robots and humans in natural ways is a primary goal of this research field. Depending on whether humans and robots are located in close proximity or not, interaction can happen in two different forms: a) remote interaction when the robot and human are not collocated, b) proximate interaction when the robot and the human are collocated \cite{goodrich2008human}.  In either form, especially in remote interaction, the user always needs some feedback from the robot side, such as the information about the state of the robot and the environment in teleoperation \cite{wang2015hermes}. To address this challenge, a number of approaches have been proposed. 
\begin{figure*}[t]
  \centering
  \includegraphics[width=\linewidth]{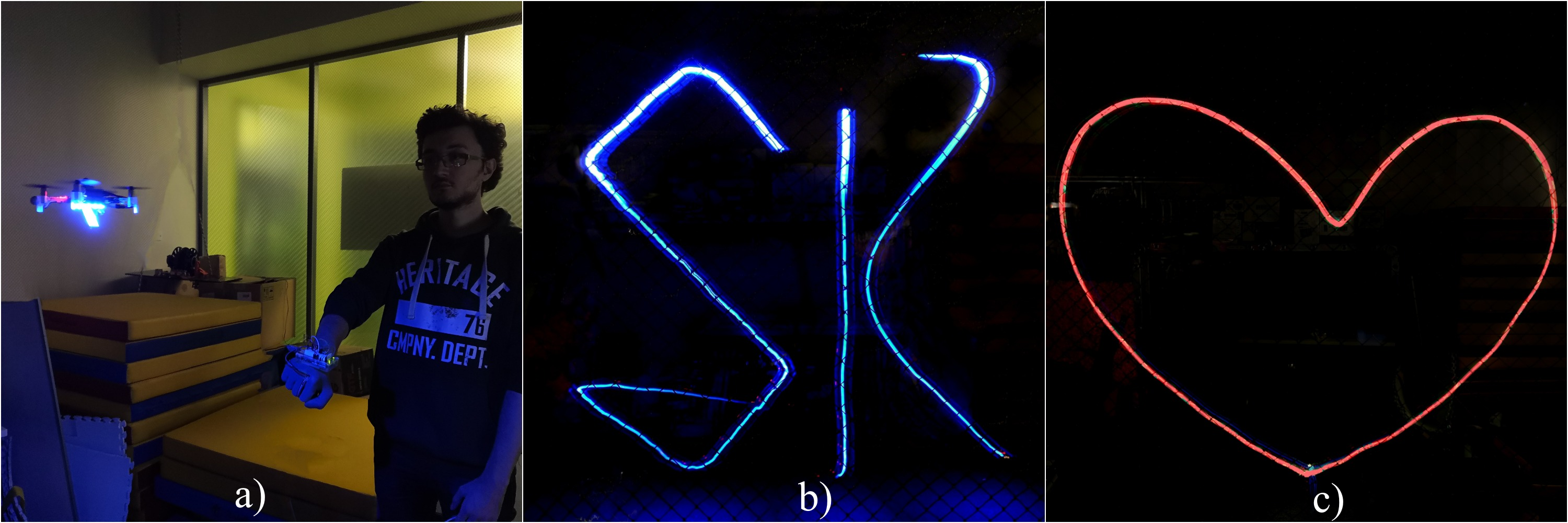}
  \caption{(a) A user draws 'S' and 'K' letters in the midair with a glove. (b) The quadcopter light-paints the letters in the midair. (c) Heart-shaped emoji light-painted by quadcopter aims at remote messaging. }~\label{fig:cats}
\end{figure*} 

\subsection{Wearable Devices for Human-Robot Interaction}
When a teleoperated robot is located in a visually constrained environment, head-mounted displays (HMDs) play an integral role in providing a visual channel to a teleoperator \cite{martins2015design}. VR applications running in HMDs are a safe and cost-effective solution that offers immersive human-robot interaction in many scenarios, e.g., robotic-surgery training, remote manipulation, and manufacturing training \cite{liu2017understanding}. In the literature, there are a plenty of papers that provide different HRI techniques through HMDs. In \cite{wang2015hermes}, the user receives visual feedback about the state of a teleoperated humanoid robot in the HMD. In another project \cite{macchini2019personalized}, the body motion of a human is translated as a control input to the robot that is simulated in the VR environment in HMD. The control of a real quadcopter, which is simulated in VR, is analyzed in \cite{erat2018drone, ibrahimov2019dronepick}. Recognition of the hand gestures is a relatively old and well-known technique \cite{fels1998glove}. Authors propose to control a quadcopter by using head position and hand gestures, which are detected by an Augmented Reality-HMD (AR-HMD) in which the trajectory of the robot is displayed \cite{walker2018communicating}. The user fixates a spatial space that is donated by a dot in AR-HMD to manipulate the quadcopter to the same space in the physical environment \cite{yuan2019human}.
\par With the advent of the tactile and force-feedback displays, haptics has gained considerable attention in the research of human-centric robotic systems encompassing teleoperation and virtual environments. When the visual channel is overloaded in the process of teleoperation, the haptic interfaces deliver the feedback information about the robot and the teleoperation side to the human somatosensory system. There are various such interfaces that have been developed. For example, a user wears FlyJacket, which renders the sensation of a flight, to control a virtual drone \cite{rognon2019perception}. 

In \cite{tsykunov2018swarmtouch, labazanova2018swarmglove, tsykunov2019swarmtouch}, the authors propose the
plan of action where impedance control is used to guide the formation of the swarm of nano-quadrotors by a human operator and the tactile patterns at the fingerprints are used to depict the dynamic state of the formation by vibrotactile feedback. Similarly, the mobile robot accurately discerns the external form, edges, movement route, velocity, and the distance to the peculiar obstacle in the robotic telepresence system \cite{tsetserukou2011belt}. Afterwards, the user who determines the movement direction and velocity of the robot by altering the his or her position through a series of movements and changes in course receives the detected information dispatched by the
tactile belt. In another proposed system \cite{scheggi2014human}, the
haptic bracelet, which delivers the tactile feedback about the feasible guidance of a group of mobile robots in the motion constrained scenarios, was developed.

\subsection{Evaluation}
Even though all the results in the literature mentioned above are significantly contributing to HRI, there are still some limitations that need to be addressed in future research. In \cite{wang2015hermes}, it is mentioned that the total delay of 175ms of the system loop is sufficient for controlling the balance of the robot. There is no experiment to verify that this time delay will be sufficient enough when the legs are added to the motion capture suit to enable stepping control. In \cite{macchini2019personalized, erat2018drone, ibrahimov2019dronepick, walker2018communicating}, it is not clear what kind of control strategy there will be if the number of agents is multiple. When there is a limited resolution depending on the distance between the operator and the quadcopter, pick-and-place command can be challenging.
Moreover, eye calibration error should not be condemned as different face structures of people may lead to different levels of errors \cite{ibrahimov2019dronepick}. There should also be some study on the psychological effects of a sudden switch from AR to VR environments. For \cite{yuan2019human}, the number of participants in the user study is considerably low to confirm the validity of the system entirely.

While there are some benefits of the haptic devices, including receiving the feedback from the points that can hardly be spotted to gain the visual feedback \cite{tsykunov2018swarmtouch}, a series of limitations cause the difficulties in applying these devices in real-life
cases widely. In fact, haptics generally requires some supplementary equipment to be configured, which in turn brings about complications in the implementation in particular applications. The low bandwidth channel in the information delivery minimizes the efficiency of the haptic feedback. For instance, the user might require to receive information
on altitude status concurrently with obstacle warning. In some very exhausting tasks, stimulus might not be sensed if a person cannot concentrate on his/her sensory input \cite{LillyTechReport}.

\subsection{DroneLight Approach}
 
 Despite the above-mentioned applications, drones are still not widely accepted by the society. Most of the drones are believed to be dangerous for human life to be used in daily life. In the meanwhile, human-controlled quadcopters usually have complicated interfaces that allow only expert users to opearte them. With this in mind, we propose a safe and easy to use DroneLight system that provides a new paradigm of HDI.  In our proposed system, as shown in Fig. 2, a user draws pre-defined patterns or letters with a glove equipped with an inertial measurement unit (IMU), in the midair. The ML algorithm matches each gesture to a unique trajectory representing the patterns or letters. In the meantime, the quadcopter, on which a light reflector and an array of controllable RGB LED array are attached, light-paints the matched patterns or letters in the real world.

  \begin{figure} [hbt!]
\centering
\includegraphics[width=0.35\textwidth]{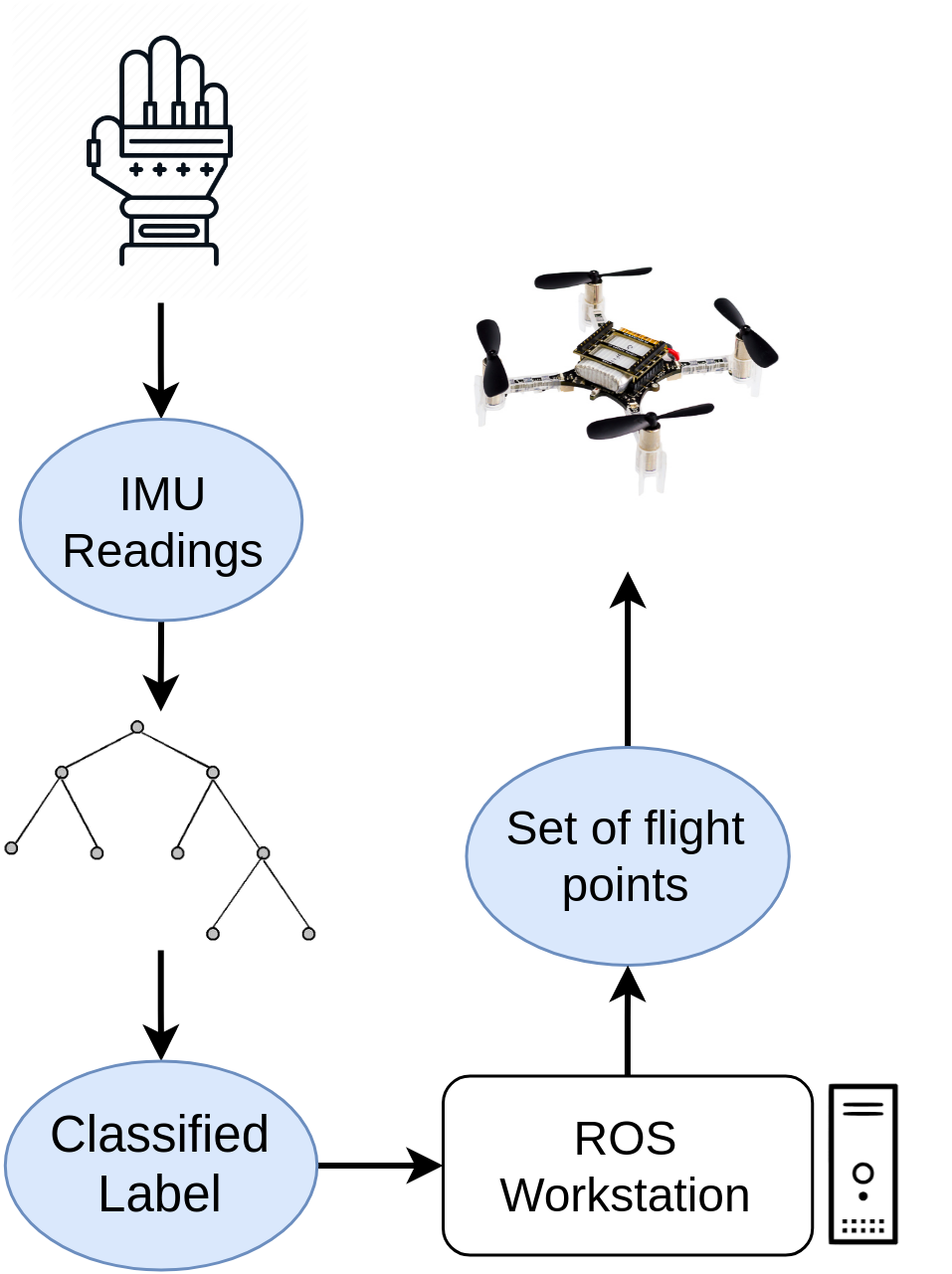}
\caption{Overview of DroneLight system.}
  \label{fig:Manipulator0}
  \vspace{-0.5em}
\end{figure}

\section{DronePick Technology}

\subsection{Drone}
The small size of 92x92x29 mm and the weight of 27 grams of Crazyflie 2.0 quadrotor make it an ideally safe platform for the HDI research. The flexibility and extendability of the platform allow researchers to attach small-sized sensory circuits to it. On the bottom of the quadcopter, one ring deck, which features twelve RGB LEDs, is attached. The quadcopter is programmed to get different color combinations throughout the flight. To achieve a more vibrant light reflection for the light-painting, a rectangular light-reflecting tape is attached to the middle of the ring deck (Fig. 3). Vicon motion capture system (Vantage V5) with 12 cameras is used to track the quadcopter within a 5x5x5 m space achieving submillimeter accuracy. The system runs on Robot Operating System (ROS) Kinetic framework.

\begin{figure} [hbt!]
\centering
\includegraphics[width=0.4\textwidth]{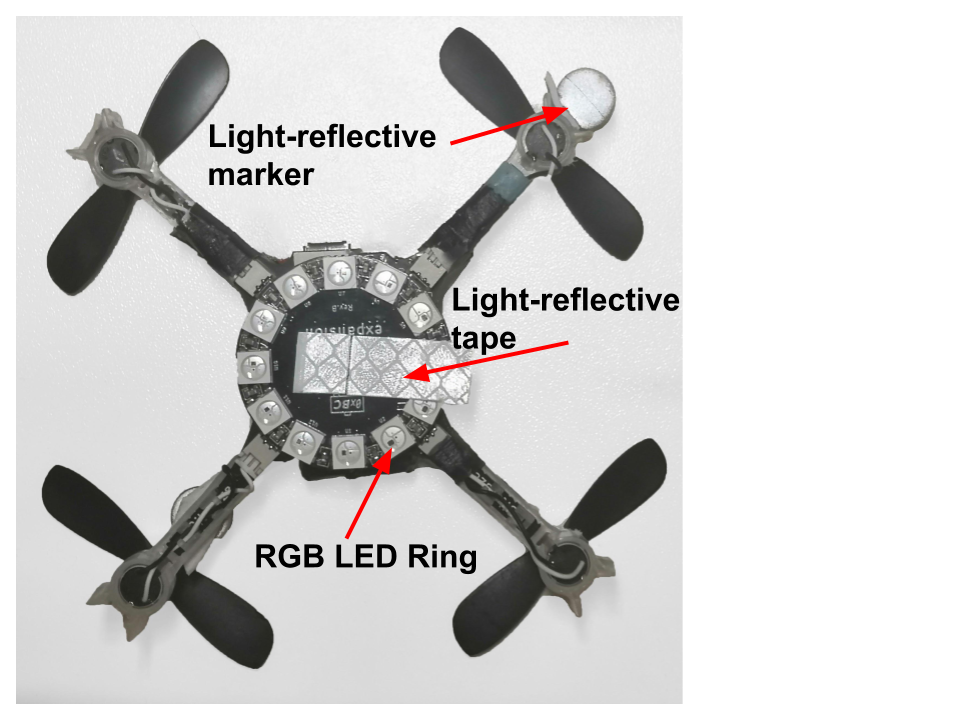}
 \caption{Configuration of the micro-quadcopter for light-painting.}
  \label{fig:Manipulator0}
  \vspace{-0.5em}
\end{figure}

\subsection{Glove}
The glove, which plays the role of input data source for DroneLight system, consists of an Arduino Uno controller, XBee module, IMU, and flex sensors, as shown in Fig. 4. The user wears it to draw one of the predefined letters in the air. Reading the values of acceleration along X, Y, and Z directions from IMU starts when the human’s hand is clasped. It provides condemning the undesired sensor values that are not part of the letters.  The clasp position of the hand is detected via the flex sensor which is located on the top of the middle finger. Sensor readings are streamed to the local PC through the XBee module.

\begin{figure} [hbt!]
\centering
\includegraphics[width=0.45\textwidth]{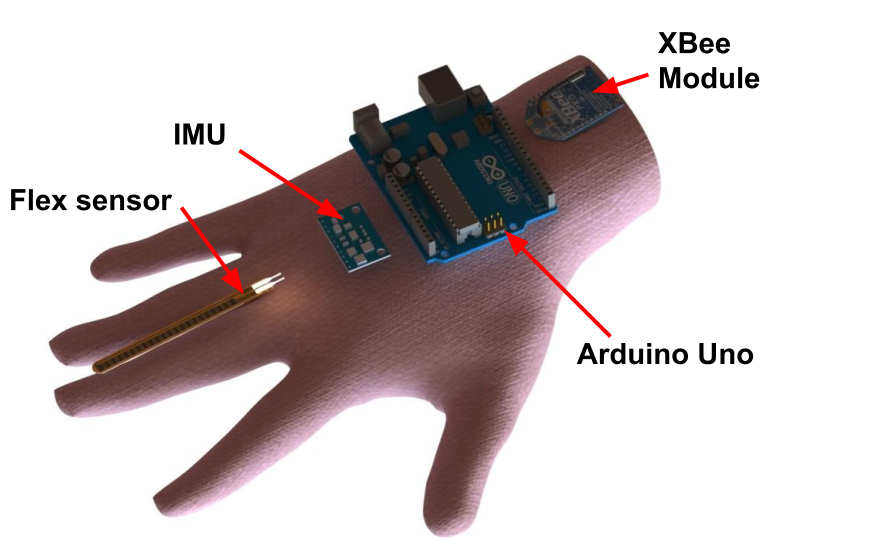}
\caption{The structure of the glove.}
  \label{fig:glove}
  \vspace{-0.5em}
\end{figure}

\section{Gesture Recognition}
Five different letters (S, K, O, L and J), which correspond to the light-paintings drawn by the quadcopter, were defined. The user draws each letter separately in the midair with the glove. For each gesture, 25 sets of data samples were collected manually. Each set consists of a sequence of 30 acceleration values in three axes. For preprocessing, we equalized the length of all input sequences by sampling within the start-end window and then smoothed the data by applying digital filter Forward-Backward Filtering to the signal. Raw and smoothed data were recorded using an accelerometer, which is  shown in Fig. 5. As a final step, all the sets were equalized by sampling within a window length. As a result, each of the input data samples is a sequence of 30-dimensional feature arrays (10 values per axis) representing the status of the glove for each time step. Example of a reduced data sample is presented in Fig. 7.

\begin{figure} [hbt!]
\centering
\includegraphics[width=0.52\textwidth]{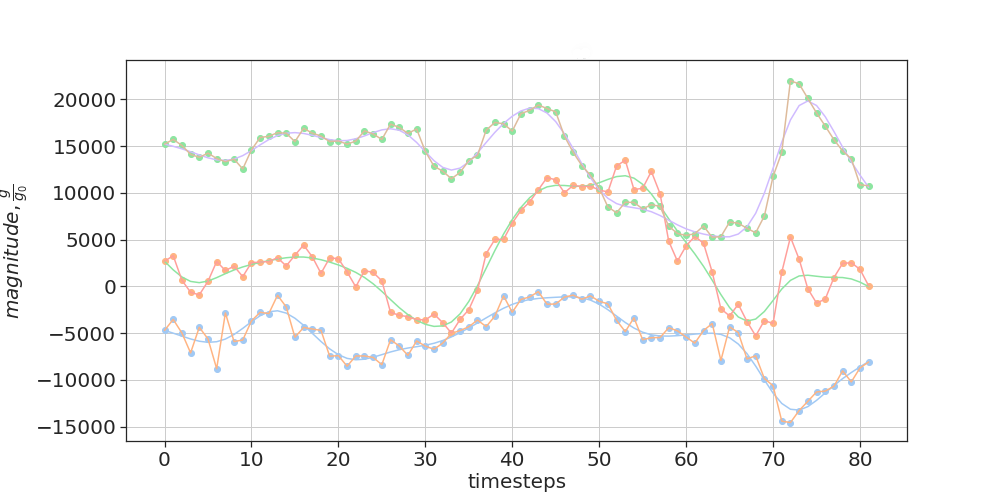}
\caption{Raw (green, orange, and blue dots accordingly) and filtered data (purple, green, and blue lines accordingly) of x, y, and z acceleration.}
  \label{fig:glove}
  \vspace{-0.5em}
\end{figure}

An algorithm based on Random Forest Classifier was developed to map five input gestures to the five flight setpoints of the quadcopter. The total dataset contains 125  sets of processed data samples. The dataset was divided into 75 and 50 sets for training and testing, respectively. Stratified K-Fold was used to keep the same distribution of classes. 

The Random Forest Classifier was evaluated with four different numbers of trees in the forest: [50, 100, 200, 300] and with four different maximum depths of the tree: [2, 3, 4, 6].  As a total, 16 Random Forest Classification models were trained. The grid search algorithm generated candidates from a grid of parameter values and recorded the accuracy scoring.

After the classifier was trained, the testing was performed to evaluate the performance. As mentioned earlier, the testing dataset has 50 samples, which are separated from the training dataset. It ensures that the classifier is evaluated with sufficient unknown information.

The results of the evaluation are listed in the confusion matrix. Among 16 trained models, the configuration of 100 trees in the forest and maximum depth of 3 showed the best result.  Since the dataset was collected based on a single person’s hand gestures, Evaluating Classifier on test set resulted in 0.98 for accuracy rate, 0.98 for precision and 0.98 for recall, which in turn showed that classification model correctly predicted 49 out of 50 input gestures (see Fig. 6). 
\begin{figure} [hbt!]
\centering
\includegraphics[width=0.4\textwidth]{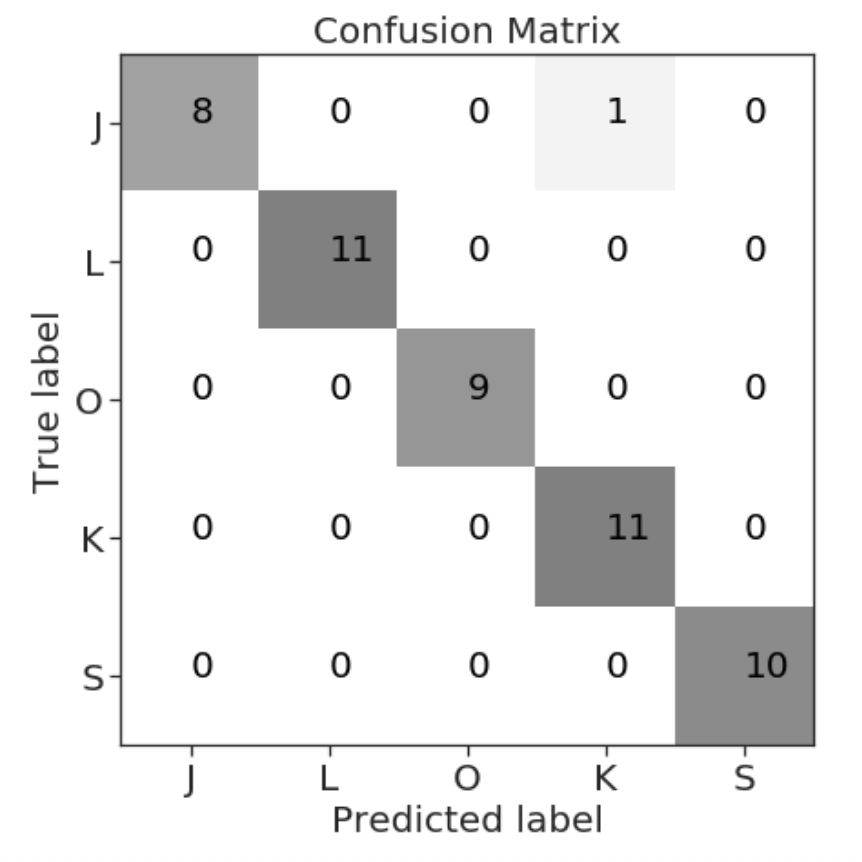}
\caption{Confusion matrix for letter recognition.}
  \label{fig:Manipulator0}
  \vspace{-0.5em}
\end{figure}

\begin{figure} [t!]
\centering
\includegraphics[width=0.53\textwidth]{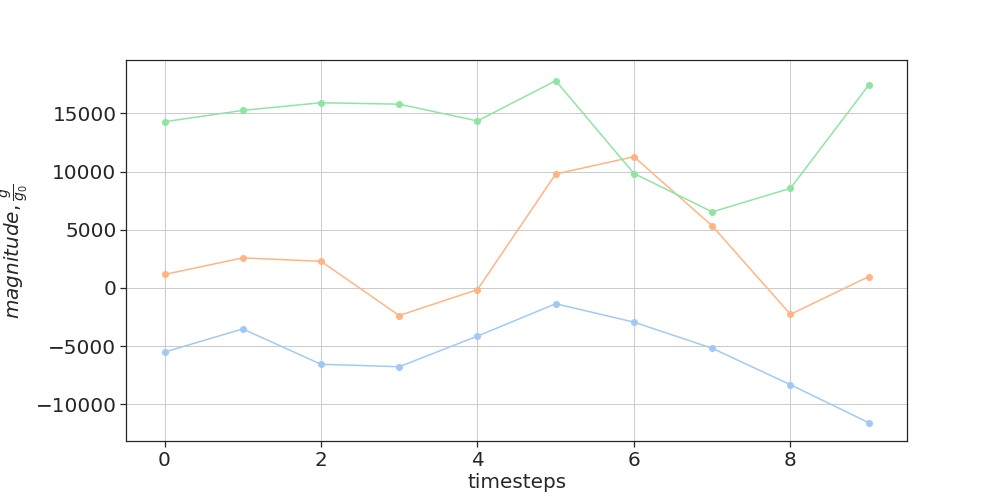}
\caption{Reduced data for x, y, and z acceleration (green, orange, and blue lines accordingly).}
  \label{fig:Manipulator0}
  \vspace{-0.5em}
\end{figure}

\section{Conclusion and Future Work}
We have proposed a novel DroneLight system that provides tangible human-drone interaction and drone-mediated communication in real life. A micro-quadcopter equipped with a RGB LED array light-paints letters or patterns in the midair based on gestures that are drawn by a user with the developed glove. For gesture recognition, a data set of five letters from a single user were collected. An ML-based algorithm was developed for gesture recognition. The recognition rate of the developed algorithm revealed high scores. The real-life implementation of the DroneLight system was also tested and verified. 

For the future, the dataset is to be enriched by the hand gesture data from different users. It will make possible for different users to use the DroneLight system. Not only additional letters but also figures (Fig.1(c)) are to be added to the dataset. A more precise and faster ML-based algorithm is to be developed.

Moreover, there are a number of possible applications that can be realized with the help of the DroneLight. It can widely be used in entertainment or photography. For example, a novel messaging system MessageDrone when partners can not only exchange messages and emoji over the distance but also enjoy the light art in starring night.  A novel rescue alarm system based on DroneLight can also be developed to substitute flare gun aimed at distress signalling. With visual trackers continuously make a video recording of the environment, the rescue team might be informed upon the detection of the message and accomplish salvation timely. In contract to flare gun, DroneLight can make it possible to deliver not only alarm but also valuable information by light-painting, i.e., GPS coordinates, what was happen, number of people, and etc.

\addtolength{\textheight}{-12cm}
\bibliographystyle{IEEEtran}
\bibliography{bib}

\end{document}